\documentclass[runningheads]{llncs}
\usepackage[T1]{fontenc}
\usepackage{algorithm}
\usepackage{algpseudocode}
\usepackage{graphicx}
\usepackage{amsfonts, amsmath, amssymb}
\usepackage{multirow}
\usepackage{xcolor}
\usepackage{float}
\usepackage{hyperref}
\usepackage{booktabs}
\usepackage{array}
\usepackage{placeins}
\begin{document}
\title{Ocular Verification for Virtual Reality}
%
%
\author{Husanpreet Singh\inst{1}\and
Robert Tran\inst{2}\orcidID{0009-0001-3390-3930}\and
Ayushree Kharel\inst{1} \and Sudipta Banerjee\inst{1}\orcidID{0009-0001-5054-2049}}
\authorrunning{}
%
\institute{University of Wyoming, USA 
\and
San Diego State University, USA\\
\email{\{hsingh1, akharel, sbanerj3\}@uwyo.edu; rtran0691@sdsu.edu}
}
\maketitle              
\begin{abstract}
Virtual reality (VR) headsets (\textit{e.g.}, Meta Quest, Apple Vision Pro) provide a seamless user experience due to their fast, frictionless, immersive interaction in a simulated environment. In such headsets, user authentication relies on biometric cues such as the iris. However, traditional iris recognition protocols may not be adequate in cases of unconstrained acquisition, which is typical of VR-based data. In this work, we examine three crucial aspects: (1) evaluating ISO/IEC 29794-6 iris quality metrics on the VRBiom dataset and analyzing their limitations, (2) addressing data-specific challenges such as off-axis gaze, non-uniform illumination, and specular reflection using generative models, and (3) performing \textit{unimodal} (iris, periocular) recognition and \textit{multimodal} score-level fusion (iris + periocular). We observe that some metrics (\textit{e.g.}, margin adequacy) fail on VR-acquired data, whereas image adjustments primarily benefit periocular recognition, and multimodal fusion lowers the EER by $\sim$11\% over unimodal iris recognition performance. Our codes are available at \url{https://github.com/rtran7530/vrsyntheye-reu.git}.

\end{abstract}
\section{Introduction}
\noindent\textbf{Motivation.} 
 The immersive potential of virtual reality (VR) and extended reality (XR) has radically transformed human-computer interaction, driving deep integration across a number of sectors, including healthcare, education, military training, and entertainment. The global market value of XR, which stood at USD 6.1 billion in 2016, surged to USD 42.83 billion by 2022 and is projected to reach a staggering USD 345.09 billion by 2030~\cite{Review}. Virtual and augmented reality head-mounted displays (HMDs) have made this a reality~\cite{MSurvey}, and their appeal relies on frictionless, embodied interaction for user convenience and privacy. But here is the caveat: \textit{how do we verify whether the legitimate user is wearing the headset?} Traditional authentication mechanisms (\textit{e.g.}, passwords, PINs, remote-controller gestures) may not be suitable for VR devices and may be susceptible to shoulder-surfing and side-channel attacks, resulting in spoofing~\cite{Review}. 

Consequently, biometric-based authentication has emerged as the natural solution; specifically, iris and periocular recognition have become the default choice for commercial HMDs. Apple's Vision Pro uses Optic ID; Meta, HTC, and Varjo devices similarly integrate eye tracking capable sensors for both interaction and authentication~\cite{Review}. However, extracting high-quality, discriminating features inside VR headsets introduces novel technical challenges for reliable authentication. Conventional iris recognition pipelines were built around cooperative, frontal, well-illuminated acquisition settings. These algorithms may fail or severely degrade when subjected to the unconstrained acquisition typical of VR headsets. HMD cameras frequently capture off-axis gaze and suffer from non-uniform infrared (IR) illumination, occlusion by eyelashes and eyelids, and severe specular reflections from the headset's illuminators ~\cite{VRBiom}. Score-level fusion of iris with periocular cues has been proposed as a partial remedy precisely because periocular texture remains informative when the iris channel degrades, but the underlying problem of assessing frame quality remains largely unaddressed for VR-specific data~\cite{Fusion}. 


\setcounter{footnote}{0}
Therefore, we need to systematically address the shortcomings of the current quality metrics and potentially revise them keeping in mind the emerging needs of VR-based deployment. Current authentication models largely overlook the necessary image-level adjustments required to counteract HMD-induced noise, leaving a significant gap in creating reliable, continuous VR authentication systems. Finally, we need to explore optimal fusion schemes that exploit the complementary features of iris and periocular cues for ocular verification in VR. \\ 


\noindent \textbf{Our Contributions.} We conduct a principled analysis of quality evaluation and ocular authentication via iris and periocular cues as follows:

    \noindent\textbf{(1)} We rigorously evaluate twelve ISO/IEC 29794-6~\footnote{\url{https://www.nist.gov/itl/iad/btg/irex-ii-iqce}} iris quality metrics on the VRBiom dataset using two frameworks: MITRE-BIQTIris and University of Notre Dame (UND). Our findings offer important insights into addressing the data-specific challenges in unconstrained VR acquisition. \\
    \textbf{(2)} We apply several image restoration and adjustment techniques using geometric correction and generative models for off-axis correction, specular reflection removal and illumination correction to improve ocular image quality. \\
    \textbf{(3)} We first perform \textit{unimodal} iris and periocular recognition, and then combine them using \textit{multimodal} score-level fusion to enhance verification performance.

\section{Related Work} 
\subsection{Iris Recognition}
Classical iris recognition progressed from Daugman's pioneering rubber-sheet unwrapping and Gabor phase demodulation~\cite{daugman}, to Wildes' Hough transform boundary fitting~\cite{Wildes}, and Ma et al.'s radial intensity variation tracking~\cite{MA} (reviewed comprehensively by Bowyer et al.~\cite{Bowyer}). These foundational frameworks matched identities by computing Hamming distances between binary IrisCodes or applying normalized correlation across spatial-frequency bands~\cite{daugman,Wildes}. Charting the transition to deep learning, Hussein et al.~\cite{DLSurvey} report that transfer-learning architectures (e.g., VGGNet, ResNet) achieve near-ceiling accuracy on controlled, well-illuminated benchmarks like CASIA and IIT Delhi. However, their recognition rates degrade substantially when confronted with cross-sensor anomalies and mobile acquisition conditions in datasets such as MICHE and UBIRISv2~\cite{DLSurvey}.

\subsection{Iris Recognition in VR Devices}
Conventional biometric models struggle with the close-range, off-angle imagery inherently captured by head-mounted displays (HMDs). To counteract this, Wang and Kumar~\cite{Egocentric} consolidate iris and periocular features via an egocentric recognition framework, while ImmerIris~\cite{ImmerIris} explores a normalization-free feature extraction paradigm alongside a massive 500,000-image VR dataset. To handle dynamic eye movements, Shapira et al.~\cite{Gaze} introduce GazeShift, an unsupervised cross-attention model that disentangles gaze from ocular appearance without labeled data. Targeting boundary delineation, Sharma et al.~\cite{VREyeSAM} fine-tune the SAM~2 foundation model with uncertainty-aware supervision to segment non-frontal, occluded irises on the VRBiom dataset~\cite{VRBiom}. Finally, because HMD-captured irises are frequently ISO-noncompliant, Boutros et al.~\cite{Fusion} demonstrate that score-level fusion with resilient periocular features significantly outperforms standalone iris verification under degraded VR capture conditions.

\subsection{Periocular Recognition}
The periocular region encompasses the skin texture, eyelids, and eyebrows surrounding the eye, and has emerged as a reliable secondary biometric modality for identity verification. Park et al.~\cite{5339068} established the feasibility of periocular recognition in the visible spectrum. This was further expanded by Woodard et al.~\cite{5597603}, who demonstrated ocular biometrics under non-ideal, Near-Infrared (NIR) illumination conditions.
Kumari and Seeja~\cite{KUMARI20221086} provide a comprehensive survey of periocular biometrics noting that periocular recognition requires minimal user cooperation compared to iris systems while offering robustness to occlusion, pose variation, and illumination changes. They report that hand-crafted feature descriptors (Local Binary Patterns (LBP)~\cite{6915725}, Binary Statistical Image Features (BSIF)~\cite{7935110}, Leung-Malik Filters~\cite{6508951}, Scale-Invariant Feature Transform (SIFT), Speeded-Up Robust Feature (SURF), Phase Intensive Local Pattern (PILP) and ensemble approaches achieve high accuracy on well-constrained benchmark datasets, but degrade substantially under mobile acquisition and cross-spectral matching conditions. 

\subsection{Periocular Recognition in VR Devices}
Integration of biometric tracking into HMDs introduces extreme, hardware-specific anomalies that challenge unconstrained periocular recognition models. Datasets such as VRBiom~\cite{VRBiom} highlight that HMD internal cameras operate at severe oblique angles and in close proximity, accompanied by intense NIR illumination. 
Baek et al.~\cite{electronics14020240} demonstrate that this proximity produces specular reflections on the cornea, causing distortions and forcing neural networks to overfit to hardware noise instead of biometric features.

\section{Proposed Approach}
\begin{figure}
    \centering
    \includegraphics[width=\textwidth]{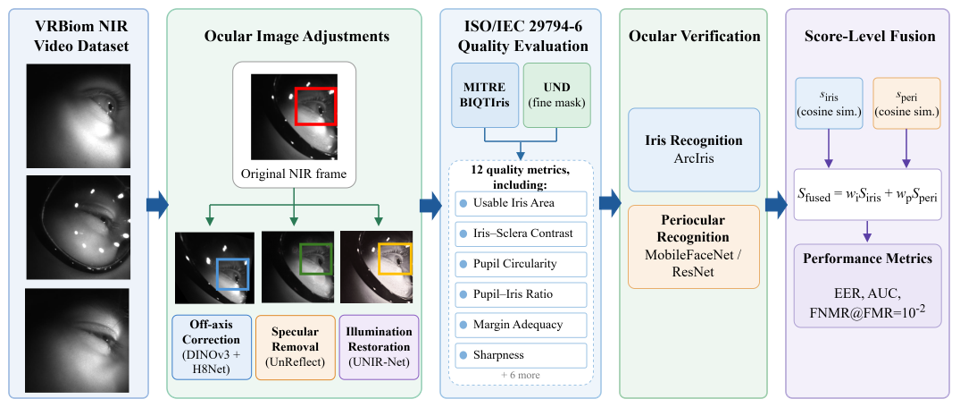}
    \caption{Overview of the proposed framework for VR-based ocular verification.}
    \label{fig:proposed_diagram}
\end{figure}

Our proposed approach (see Fig.~\ref{fig:proposed_diagram}) consists of three distinct processing modules. Firstly, we perform a set of \textbf{Ocular Image Adjustments}, namely, off-axis gaze correction, specular reflection removal and uneven or non-uniform illumination adjustment using machine learning and generative models for overall quality enhancement; see Algorithm~\ref{alg:image_correction}. Secondly, we perform a \textit{first-of-its-kind} comprehensive analysis of \textbf{ISO/IEC 29794-6 Quality Evaluation} in VR-based ocular images using two frameworks developed independently by MITRE and UND; see Algorithm~\ref{alg:quality_eval}. Thirdly, we perform \textbf{Ocular Verification and Multimodal Fusion} via iris recognition (using ArcIris), periocular recognition (using MobileFaceNet), and finally a weighted multimodal iris and periocular fusion at score level for ocular verification; see Algorithm~\ref{alg:verification_and_fusion}. 

\subsection{Ocular Image Adjustments}
We address off-axis perspectives, hardware-induced specular reflections, and uneven or non-uniform illumination in VR data using three operations. 
\textbf{Off-axis Correction (DINOv3+H8Net):}
Algorithm \ref{alg:image_correction} details the pipeline that uses geometric off-angle correction ($c_o$) to mathematically estimate and mitigate the perspective distortion inherent to off-axis VR headset sensor arrays. The DINOv3 foundation model is utilized to extract dense spatial features from the ocular image ($f_{curr}$). These features are subsequently passed to the H8Net regression head to estimate an 8-degree-of-freedom (8-DoF) projective homography matrix $H$. This matrix is then applied via a forward perspective warp to reproject the raw, off-axis elliptical iris back into a canonical frontal circular geometry, restoring spatial proportions prior to the polar unwrapping phase.

\textbf{Specularity Removal (UnReflect):}
Specular glare and reflections introduced by the VR headset's internal near-infrared (NIR) illuminators mask the fine-grained details of the iris pattern, necessitate advanced reflection-removal techniques. To remove this, we adopt the UnReflectAnything ~\cite{Rota_2026_CVPR} framework, an RGB-only single-image highlight removal pipeline. We use a frozen vision transformer backbone (DINOv3)~\cite{simeoni2025dinov3} to extract multi-scale feature representations from the periocular data, while the inpainting module fills in the corrupted regions to reconstruct a reflection-free image. We hypothesize that we can leverage the underlying reflection removal mechanism on near-infrared data. 

\textbf{Non-Uniform Illumination Restoration (UNIR-Net):}
Images captured by VR headsets exhibit non-uniform illumination due to near-infrared illuminators and variations in lighting across the eye region, resulting in uneven brightness and reduced visibility of the iris. To address this, we employ UNIR-Net~\cite{P_rez_Zarate_2025}, an illumination restoration network. The network combines illumination enhancement, attention mechanisms, visual refinement, and contrast correction to improve image quality and visibility. To satisfy the 3-channel RGB input requirements of the pre-trained backbones (e.g., DINOv3), we replicate the single-channel NIR frames across three channels before inference.

\begin{algorithm}[!t]
\caption{Ocular Image Enhancement and Adjustment}
\label{alg:image_correction}
\begin{algorithmic}[1]
\fontsize{8.5pt}{8.5pt}\selectfont
\State \textbf{Input:} Grayscale Frame Dataset $F$, Network Models $\{M_{UR}, M_{UNIR}, M_{DINO}, M_{H8}\}$, Configuration $C = \{c_{u}, c_{n}, c_{o}\}$ \Comment{$c_u$: UnReflect, $c_n$: UNIR-NET, $c_o$: Off-axis Correction}
\State \textbf{Output:} Output Dataset $F_{out}$
\vspace{1.5mm}

\Function{EnhanceDataset}{$F$, Models, $C$}
    \State $F_{out} \gets \emptyset$
    
    \For{\textbf{each} frame $f \in F$}
        \State $f_{curr} \gets f$
        
        \If{$c_{u}$ is \textbf{True}}
            \State $f_{curr} \gets \Call{ApplyUnReflect}{f_{curr}, M_{UR}}$ \Comment{Mitigate hardware specular glare}
        \EndIf

        \If{$c_{n}$ is \textbf{True}}
            \State $f_{curr} \gets \Call{ApplyUNIR-NET}{f_{curr}, M_{UNIR}}$ \Comment{Illumination correction}
        \EndIf

        \If{$c_{o}$ is \textbf{True}}
            \State $H \gets \Call{HomographyMatrix}{f_{curr}, M_{DINO}, M_{H8}}$ \Comment{Extract 8 spatial parameters, Build $3 \times 3$ matrix ($H_{3,3} = 1.0$)}
            \State $f_{curr} \gets \Call{WarpPerspective}{f_{curr}, H}$ \Comment{Apply geometric flattening}
        \EndIf
        
        \State $F_{out} \gets F_{out} \cup \{f_{curr}\}$
    \EndFor
    
    \State \textbf{return} $F_{out}$
\EndFunction
\end{algorithmic}
\end{algorithm}

\subsection{ISO Quality Evaluation}
We perform a comprehensive statistical analysis of different ISO/IEC quality metrics based on implementations of two frameworks developed by MITRE BIQTIris and UND. Unlike traditional biometric datasets that use controlled, frontal acquisition, VR devices present a unique set of constraints that \textit{necessitate a re-evaluation of established quality benchmarks}. Our primary objective is to empirically derive the statistical distributions ($\mu \pm \sigma$) of standard ISO/IEC 29794-6 metrics across our dataset (see Table~\ref{tab:iso_metrics_master}), thereby establishing a normative baseline for future VR-based biometric studies. We utilize the procedure detailed in Algorithm \ref{alg:quality_eval} to systematically process our dataset, providing researchers with the statistical spread of current VR imagery.

\subsection{Ocular Verification and Fusion}
We first perform ocular verification using \textit{unimodal} schemes: (i) iris recognition using ArcIris, a deep learning-based model submitted by UND to the NIST IREX X evaluation, (ii) periocular recognition using lightweight MobileFaceNet and ResNet models. Secondly, to overcome the inherent degradations of unimodal iris recognition, we performed a \textit{multimodal weighted score-level fusion} scheme combining iris and periocular similarities with varying weights.
\begin{equation}
    S_{fused} = w_i \times S_{iris} + w_p \times S_{peri}
\end{equation}
where $S_{iris}$ and $S_{peri}$ denote the iris and periocular similarity scores, and $w_i$ and $w_p$ are their respective weights.

\begin{table*}
    \centering
    \footnotesize
    \caption{Mathematical definitions of standard ISO/IEC 29794-6 quality metrics.}
    \label{tab:iso_metrics_master}
    \resizebox{\textwidth}{!}{%
    \begin{tabular}{>{\raggedright\arraybackslash}p{0.19\textwidth} >{\raggedright\arraybackslash}p{0.26\textwidth} >{\raggedright\arraybackslash}p{0.40\textwidth} >{\raggedright\arraybackslash}p{0.18\textwidth}}
        \toprule
        \textbf{Metric} & \textbf{Description} & \textbf{Formulation} & \textbf{Variable Definitions} \\
        \midrule

        \textbf{Usable Iris Area} &
        Percentage of the iris area not occluded by eyelids, eyelashes, or specular highlights. &
        $\text{UIA} = \left(1 - \frac{N_{occ}}{N_{iris}}\right) \times 100$ &
        $N_{iris}$: Iris ring pixels \newline $N_{occ}$: Occluded pixels \\
        \addlinespace[0.8ex]

        \textbf{Iris-Sclera Contrast} &
        Contrast along the outer iris-sclera boundary to prevent radial segmentation failure. &
        $\text{ISC} = \frac{|SV - IV_{isc}|}{SV + IV_{isc} - 2PV} \times 100$ \newline
        \scriptsize{(0 if $PV\!\geq\! IV_{isc}$ or $PV\!\geq\! SV$)} &
        $SV$: Sclera median \newline $IV_{isc}$: Iris median \newline $PV$: Pupil median \\
        \addlinespace[0.8ex]

        \textbf{Iris-Pupil Contrast} &
        Contrast across the inner pupil boundary for precise inner-circle localization. &
        $\text{IPC} = \frac{WR}{0.75 + WR} \times 100$ \newline \scriptsize{where $WR = \frac{|IV_{ipc} - PV|}{20 + PV}$} &
        $IV_{ipc}$: Inner iris median \newline $WR$: Weber Ratio \\
        \addlinespace[0.8ex]

        \textbf{Pupil Circularity} &
        Degree to which segmented pupillary boundary approximates a true circle via Fourier series. &
        $\text{PBC} = \max\left(0, 100 - \frac{1}{N} \sum_{k=1}^{M-1} \|C_k\|^2\right)$ &
        $C_k$: Fourier coefficients of radial distances \\
        \addlinespace[0.8ex]

        \textbf{Grayscale Utilization} &
        Dynamic range and scattering of pixel intensities evaluated via histogram entropy. &
        $\text{GSU} = -\sum_{i=0}^{255} p_i \log_2 p_i$ &
        $p_i$: Intensity probability density \\
        \addlinespace[0.8ex]

        \textbf{Iris Radius}$^{*}$ &
        Radius of the circle approximating the segmented outer iris-sclera boundary, in pixels. &
        $R_{iris} = \sqrt{(X_b - X_i)^2 + (Y_b - Y_i)^2}$ &
        $(X_i, Y_i)$: Iris center \newline $(X_b, Y_b)$: Outer boundary \\
        \addlinespace[0.8ex]

        \textbf{Pupil-Iris Ratio} &
        Degree of pupil dilation relative to total iris size. Extreme dilation distorts texture. &
        $\text{PIR} = \frac{R_{pupil}}{R_{iris}} \times 100$ &
        $R_{pupil}$: Pupil radius \newline $R_{iris}$: Iris radius \\
        \addlinespace[0.8ex]

        \textbf{Concentricity} &
        Spatial alignment between pupil and iris centers. Low scores indicate severe off-axis gaze. &
        $\text{IPC}_o = \max\left(0, 1 - \frac{d}{R_{iris}}\right) \times 100$ \newline \scriptsize{where $d = \sqrt{(X_p - X_i)^2 + (Y_p - Y_i)^2}$} &
        $(X_p, Y_p)$: Pupil center 
        $(X_i, Y_i)$: Iris center \newline
        $d$: Euclidean distance \\
        \addlinespace[0.8ex]

        \textbf{Margin Adequacy} &
        Iris centralization relative to frame edges. Highly sensitive to off-axis VR camera angles. &
        $\text{MA}_{deq} = 100 \cdot \min\{LM', RM', UM', DM'\}$ \newline
        \scriptsize{$LM'\!=\!\text{clip}(LM/0.6,0,1)$, similarly for others} &
        $LM, RM$: Horiz. margins $/0.6$ \newline $UM, DM$: Vert. margins $/0.2$ \\
        \addlinespace[0.8ex]

        \textbf{Sharpness} &
        Focus clarity of fine textures, evaluated via Laplacian-of-Gaussian (LoG) signal power. &
        $\text{SHP} = 100 \cdot \frac{P^2}{P^2 + c^2}$ \newline \scriptsize{where $P = \frac{1}{w_F \cdot h_F} \sum I_F(x,y)^2$} &
        $I_F$: LoG convolved image \newline $P$: Signal power; $c$: Constant. \\

        \bottomrule
    \end{tabular}%
    }
    \\[0.5ex]
    \scriptsize{$^{*}$Formula inferred from the standard's description; not explicitly stated as an equation in the source metric definitions~\cite{Brazil}. Excludes motion blur and overall quality.}
\end{table*}

\begin{algorithm}[t]
\caption{ISO Metric Quality Evaluation and Statistical Analysis}
\label{alg:quality_eval}
\begin{algorithmic}[1]
\fontsize{8.5pt}{8.5pt}\selectfont
\State \textbf{Input:} Evaluated Frame Dataset $F \in \{F_{orig}, F_{OAC}, F_{UR}, F_{UNIR}\}$, BIQT engine $E_{BIQT}$, Set of Target ISO Metrics $\mathcal{Q}$
\State \textbf{Output:} Dataset Statistical Summary $T_{stats}$ (containing $\mu$ and $\sigma$ for all $q \in \mathcal{Q}$)
\vspace{1.5mm}

\Function{EvaluateQuality}{$F, E_{BIQT}, \mathcal{Q}$}
    \State $T_{stats} \gets \emptyset$
    
    \For{\textbf{each} $q \in \mathcal{Q}$}
        \State $V_q \gets \emptyset$ \Comment{Initialize empty sequence for metric values}
    \EndFor
    \For{\textbf{each} frame $f \in F$}
        
        \State $R \gets \Call{RunBiqt}{f, E_{BIQT}}$ \Comment{Generate JSON quality report}
        \For{\textbf{each} $q \in \mathcal{Q}$}
        
            \State $v \gets \Call{ParseMetric}{R, q}$ \Comment{Extract scalar quality value}
            \State $V_q \gets V_q \cup \{v\}$
        \EndFor
    \EndFor
    
    \For{\textbf{each} $q \in \mathcal{Q}$}
        \State $\mu_q \gets \Call{Mean}{V_q}$
        \State $\sigma_q \gets \Call{StandardDeviation}{V_q}$ 
        
        \State $T_{stats} \gets T_{stats} \cup \{(q, \mu_q, \sigma_q)\}$ \Comment{Compile final distributions}
    \EndFor
    
    \State \textbf{return} $T_{stats}$
\EndFunction
\end{algorithmic}
\end{algorithm}

\begin{algorithm}[!t]
\caption{Ocular Verification and Multimodal Score-Level Fusion}
\label{alg:verification_and_fusion}
\begin{algorithmic}[1]
\fontsize{8.5pt}{6.5pt}\selectfont
\State \textbf{Input:} Frame variants $F$, Video variants $\mathcal{V}$, Models $\mathcal{M} = \mathcal{M}_{iris} \cup \mathcal{M}_{peri}$ (where $\mathcal{M}_{iris} = \{M_{iris}\}$), Condition sets $V_{iris}, V_{peri}$, Periocular pairs $D_{peri}^{\text{orig}}$, Fusion weights $W$
\State \textbf{Output:} Unimodal \& Fusion metrics $T_{eval}$ (EER, FMR, FNMR, $d'$, AUC)
\Statex
\State \textbf{\textit{Part 1: Unimodal Verification (Iris \& Periocular)}}
\Function{UnimodalVerification}{$F, \mathcal{V}, \mathcal{M}, V_{iris}, V_{peri}$}
    \State $\mathcal{I}_{iris} \gets \{i \mid \textsc{CheckQuality}(M_{iris}, f_i^{(\text{OAC})})\}$ \quad $\triangleright$ \text{Indices of OAC-quality survivors}
    \State $T_{eval} \gets \emptyset$
    \For{\textbf{each} modality $X \in \{\text{iris}, \text{peri}\}$, variant $v \in V_X$, and model $m \in \mathcal{M}_X$}
        \State $F_{\text{eval}} \gets \{f_i^{(v)} : i \in \mathcal{I}_{iris}\}$ \textbf{if} $X = \text{iris}$ \textbf{else} $\textsc{RandomSample}(\mathcal{V}^{(v)}, k=3)$
        \State $S_{X, v, m} \gets \textsc{ComputeScores}(m, F_{\text{eval}})$
        \State $T_{eval} \gets T_{eval} \cup \{(X, v, m, \textsc{ComputeStats}(S_{X, v, m}))\}$
    \EndFor
    \State \textbf{return} $T_{eval}$
\EndFunction \\

\State \textbf{\textit{Part 2: Multimodal Score-Level Fusion}}
\Function{ScoreLevelFusion}{$D_{peri}^{\text{orig}}, M_{iris}, W$}
    \State $S_{iris}, S_{peri}, Y \gets \emptyset, \emptyset, \emptyset$ \quad $\triangleright$ \text{Y represents ground truth labels (0/1)}
    \For{\textbf{each} pair $(I_p, I_r, y, s_{peri}) \in D_{peri}^{\text{orig}}$} \quad $\triangleright$ \text{Process 1.8M exact pairs}
    
               \If{$\neg\Call{CheckQuality}{M_{iris}, I_p}
    \lor \neg\Call{CheckQuality}{M_{iris}, I_r}$}
    \State $s_{iris} \gets 0.0$
\Else
    \State $s_{iris} \gets
    \Call{ComputeSim}{M_{iris}, I_p, I_r}$
\EndIf
        
        \If{$s_{iris} > 0.0$ \textbf{and} $s_{peri} > 0.0$} \quad $\triangleright$ \text{Intersection filter \& FTE penalty}
           \State Append $s_{iris}$ to $S_{iris}$,
$s_{peri}$ to $S_{peri}$, and $y$ to $Y$
        \EndIf
    \EndFor
    
    \State $T_{eval} \gets \emptyset$
    
    \For{\textbf{each} $(w_i, w_p) \in W$}
        \State $T_{eval} \gets T_{eval} \cup \{(w_i, w_p, \textsc{ComputeFusionStats}(w_i S_{iris} + w_p S_{peri}, Y))\}$    
    \EndFor
    \State \textbf{return} $T_{eval}$
\EndFunction
\end{algorithmic}
\end{algorithm}

\section{Experiments and Analysis}
\subsection{Dataset}
We utilize the publicly available VRBiom dataset introduced by Kotwal et al.~\cite{VRBiom}. The data is acquired using inward-facing near-infrared (NIR) tracking cameras of a Meta Quest Pro headset at a spatial resolution of $400 \times 400$ pixels. It comprises 900 bonafide video recordings across 25 diverse subjects. To encompass real-world behavioral and hardware variations, such as, off-axis camera perspectives and three dynamic gaze conditions (steady gaze, moving gaze, and partially closed eyes) as well as two eyewear configurations (with and without glasses). The dataset also includes presentation attacks but we have confined our experiments to the bonafide data. We extract frames from each video, obtaining $\sim 23\text{K}$ frames per subject on average ($=72 (\text{fps}) \times  9 (\text{secs}) \times 3 (\text{gaze conditions}) \times 3 (\text{trials for each gaze condition}) \times 2 (\text{with/without glasses}) \times 2 (\text{left and right eye})$). For periocular recognition experiments, we randomly sampled 3 frames per video resulting in $\sim$110 frames per subject.

\subsection{Metrics}
We report the results in terms of (1) \textbf{FNMR@FMR$=10^{-2}$}: False Non-Match Rate (FNMR) at a specific False Match Rate (FMR), lower is better;
(2) \textbf{Equal Error Rate (EER)}: denotes the operating point at which FMR and FNMR are equal, lower is better; (3)
\textbf{Area Under the Curve (AUC)}: denotes the area under the receiver operating characteristics (ROC) curve, higher is better; (4) \textbf{d-prime (d')}: measures the distinguishability between genuine and imposter score distributions; a measure from signal detection theory. It is computed as $\displaystyle \frac{{\mu_{gen}-\mu_{imp}}}{\sqrt{0.5\times (\sigma_{gen}^2+\sigma_{imp}^2)}}$ (assuming similarity scores), higher is better. 

\subsection{Implementation Details}
We used the MITRE BIQTIris framework\footnote{\url{https://github.com/mitre/biqt-iris}} and the internal implementation developed by the University of Notre Dame (UND) with \textit{fine} segmentation mask for ISO/IEC 29794-6 metric evaluation. We utilized the official open-source implementation of UnReflect~\cite{Rota_2026_CVPR}, UNIR-Net~\cite{P_rez_Zarate_2025}, and periocular recognition~\cite{Periocular}. We used ArcIris~\cite{CVRL} based on \texttt{openiris-cvrl} implementation~\footnote{\url{https://github.com/CVRL/OpenSourceIrisRecognition}}. All inference and evaluation pipelines were executed using either an NVIDIA L40S or an H100 GPU. Error curves were plotted using open-source PyEER library\footnote{\url{https://github.com/manuelaguadomtz/pyeer/tree/master}}.

\subsection{Results}

\begin{figure}
    \centering
    \includegraphics[width=0.95\linewidth]{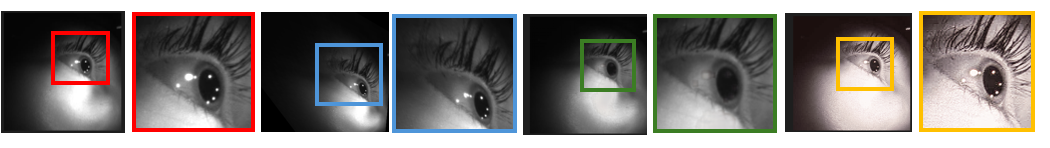}
    \includegraphics[width=0.95\linewidth]{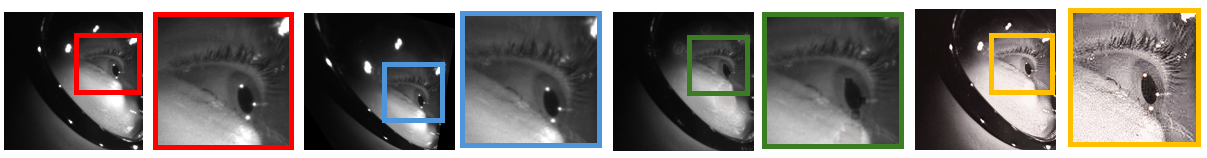}
    \caption{Examples of ocular image adjustment operations. $\textcolor{red}{\square}$: original frame; $\textcolor{blue}{\square}$: output after off-axis correction via DINOv3+H8Net; $\textcolor{green}{\square}$: output after specular reflection removal via UnReflect; $\textcolor{yellow}{\square}$: output after uneven illumination correction via UNIR-Net. Zoomed-in views of the iris regions indicate effects after processing operations.}
    \label{fig:illumadj}
\end{figure}

\subsubsection{Findings of Ocular Image Adjustment}
Fig.~\ref{fig:illumadj} illustrates that off-axis correction shows limited success, while UnReflect  can remove specular reflection and UNIR-Net can mitigate non-uniform illumination. However, since both UnReflect and UNIR-Net were trained on RGB images rather than NIR data, the adjustment operations may subtly alter the underlying fine-grained iris texture. We speculate that fine-tuning or domain adapting these generative models on NIR data will help prevent the texture
hallucination.

\subsubsection{Findings of ISO Quality Metrics}
\label{sec:ISO_eval}
Table~\ref{tab:ISO_values} reports $\mu \pm \sigma$ of each metric from both frameworks on the original, OAC, UnReflect, and UNIR-Net images. The results indicate that the VR-based data may not satisfy currently recommended values for most of the metrics. Our analysis further reveals significant divergence between algorithmic implementations based on segmentation routines and handling of invalid cases between the two frameworks. For example, while MITRE BIQTIris evaluates Usable Iris Area as $97.59 \pm 7.63$, the UND implementation produces $54.29 \pm 22.75$, with scores collapsing to $43.26$ following illumination restoration. Furthermore, the traditional ISO/IEC metric framework for margin adequacy and concentricity does not work on VR-native imagery. Thus, we recommend that ISO/IEC framework should accommodate diverse acquisition challenges present in VR devices and revise thresholds or metrics for reliable validation of VR-based image quality. OAC yields overall poor performance across both frameworks compared to the original frames due to the geometric warping.

We further investigate why the Margin Adequacy ($MA_{deq}$) metric (see Table \ref{tab:iso_metrics_master} for details) produces values substantially below the recommended threshold of 80 or greater. In our experiments, the calculated ISO metric value (using MITRE) was $4.38 \pm 16.37$. This deviation is primarily due to the off-axis orientation compared to the near-frontal, on-axis captures commonly used in datasets such as BIQT. The final $MA_{deq}$ score is determined by the \textbf{smallest of the four margins}, meaning the metric is constrained by the narrowest distance between the iris and the frame boundary. Therefore, the low mean score is likely attributable to the off-axis camera setup rather than evidence of poor image quality.

\subsubsection{Findings of Ocular Verification and Fusion}
Table~\ref{tab2} reports verification performance on near-infrared periocular frames extracted from the VRBiom~\cite{VRBiom} dataset in three settings: original frames, UnReflect, and UNIR-Net-processed frames, averaged over the three model folds. We formed the genuine pairs from distinct capture sequences of the same subject and eye, and impostor pairs from different subjects of the same eye side. Among the pre-processing methods, UnReflect lowers the EER and raises the AUC for every model. We also evaluated on off-axis-corrected (OAC) images, which produced the lowest performance, probably due to the padding that interferes with periocular recognition.




\begin{table}[htbp]
\centering
\renewcommand{\arraystretch}{1.7} 
\setlength{\tabcolsep}{8pt}       
\caption{Mean and standard deviation ($\mu \pm \sigma$) of ISO/IEC iris quality metrics for the VRBiom dataset under three processing conditions, evaluated using \textbf{MITRE BIQTIris} and \textbf{University of Notre Dame} frameworks.}
\label{ISO_Combined}
\makebox[\textwidth][c]{\scalebox{0.78}{\rotatebox{90}{
\begin{tabular}{|>{\raggedright\arraybackslash}p{3.5cm}|c|c|c|c|c|c|c|c|}
\hline
\multicolumn{1}{|c|}{\multirow{2}{*}{\textbf{Metric}}} & \multicolumn{4}{c|}{\textbf{MITRE BIQTIris ($\mu \pm \sigma$)}} & \multicolumn{4}{c|}{\textbf{University of Notre Dame ($\mu \pm \sigma$)}} \\ \cline{2-9} 
\multicolumn{1}{|c|}{}                                 & \textbf{Original} & \textbf{OAC} & \textbf{UnReflect} & \textbf{UNIR-Net} & \textbf{Original} & \textbf{OAC} & \textbf{UnReflect} & \textbf{UNIR-Net} \\ \hline \hline
USABLE\_\allowbreak IRIS\_\allowbreak AREA             & $97.59 \pm 7.63$  & $97.02 \pm 5.48$ & $98.68 \pm 4.88$   & $98.09 \pm 5.58$  & $54.29 \pm 22.75$ & $46.39 \pm 28.94$ & $56.33 \pm 21.92$  & $43.26 \pm 26.22$ \\
IRIS\_\allowbreak SCLERA\_\allowbreak CONTRAST         & $44.60 \pm 27.45$ & $40.11 \pm 37.49$ & $43.26 \pm 25.03$  & $61.60 \pm 22.61$ & $23.76 \pm 14.97$ & $22.25 \pm 16.13$ & $25.24 \pm 15.25$  & $34.65 \pm 32.30$ \\
IRIS\_\allowbreak PUPIL\_\allowbreak CONTRAST          & $58.65 \pm 23.48$ & $46.28 \pm 20.92$ & $60.47 \pm 22.61$  & $57.28 \pm 22.42$ & $34.78 \pm 10.01$ & $33.02 \pm 7.79$ & $35.07 \pm 10.41$  & $36.20 \pm 19.89$ \\
PUPIL\_\allowbreak BOUNDARY\_\allowbreak CIRCULARITY   & $88.04 \pm 17.88$ & $86.54 \pm 17.54$ & $88.58 \pm 16.06$  & $88.04 \pm 15.81$ & $74.59 \pm 35.03$ & $83.87 \pm 31.68$ & $75.09 \pm 34.35$  & $76.78 \pm 36.55$ \\
GREY\_\allowbreak SCALE\_\allowbreak UTILIZATION       & $6.84 \pm 0.63$   & $6.23 \pm 0.64$ & $6.86 \pm 0.58$    & $7.26 \pm 0.55$   & $98.06 \pm 12.25$ & $76.95 \pm 38.20$ & $98.79 \pm 9.29$   & $70.34 \pm 40.41$ \\
IRIS\_\allowbreak RADIUS                               & $145.27 \pm 42.39$ & $156.03 \pm 52.64$ & $149.85 \pm 41.39$ & $118.47 \pm 24.08$ & $4.99 \pm 10.77$  & $21.73 \pm 38.61$ & $4.03 \pm 7.81$    & $10.24 \pm 15.74$ \\
PUPIL\_\allowbreak IRIS\_\allowbreak RATIO             & $46.52 \pm 15.32$ & $39.08 \pm 15.55$ & $47.25 \pm 14.99$  & $46.67 \pm 13.92$ & $15.89 \pm 26.51$ & $7.89 \pm 16.47$ & $23.22 \pm 30.96$  & $14.56 \pm 24.94$ \\
IRIS\_\allowbreak PUPIL\_\allowbreak CONCENTRICITY     & $75.41 \pm 9.70$  & $75.29 \pm 10.17$ & $76.65 \pm 9.13$   & $75.14 \pm 9.18$  & $4.83 \pm 1.38$   & $3.01 \pm 2.25$ & $5.08 \pm 1.05$    & $5.79 \pm 1.51$   \\
MARGIN\_\allowbreak ADEQUACY                           & $4.38 \pm 16.37$  & $3.15 \pm 14.76$ & $3.53 \pm 14.69$   & $3.97 \pm 15.31$  & $96.76 \pm 9.65$  & $97.63 \pm 8.98$ & $96.62 \pm 9.96$   & $96.76 \pm 10.56$ \\
SHARPNESS                                              & $7.98 \pm 9.58$  & $4.07 \pm 7.89$ & $0.95 \pm 1.12$    & $38.34 \pm 18.25$ & $22.69 \pm 20.21$ & $20.00 \pm 17.57$ & $3.08 \pm 3.24$    & $64.21 \pm 23.88$ \\
MOTION\_\allowbreak BLUR                               & NA      & NA          &   NA               &  NA               & $1.45 \pm 0.35$   & $1.50 \pm 0.37$ & $1.60 \pm 0.35$    & $1.56 \pm 0.38$   \\
OVERALL\_\allowbreak QUALITY                           & $1.12 \pm 10.51$  & $0.71 \pm 8.11$ & $0.27 \pm 2.83$    & $0.92 \pm 9.54$   & $1.24 \pm 3.05$  & $0.81 \pm 2.25$  & $0.18 \pm 0.54$    & $2.82 \pm 6.72$   \\ \hline
\end{tabular}}}}
\label{tab:ISO_values}
\end{table}

\begin{figure*}[]
    \centering
    \includegraphics[width=0.9\textwidth]{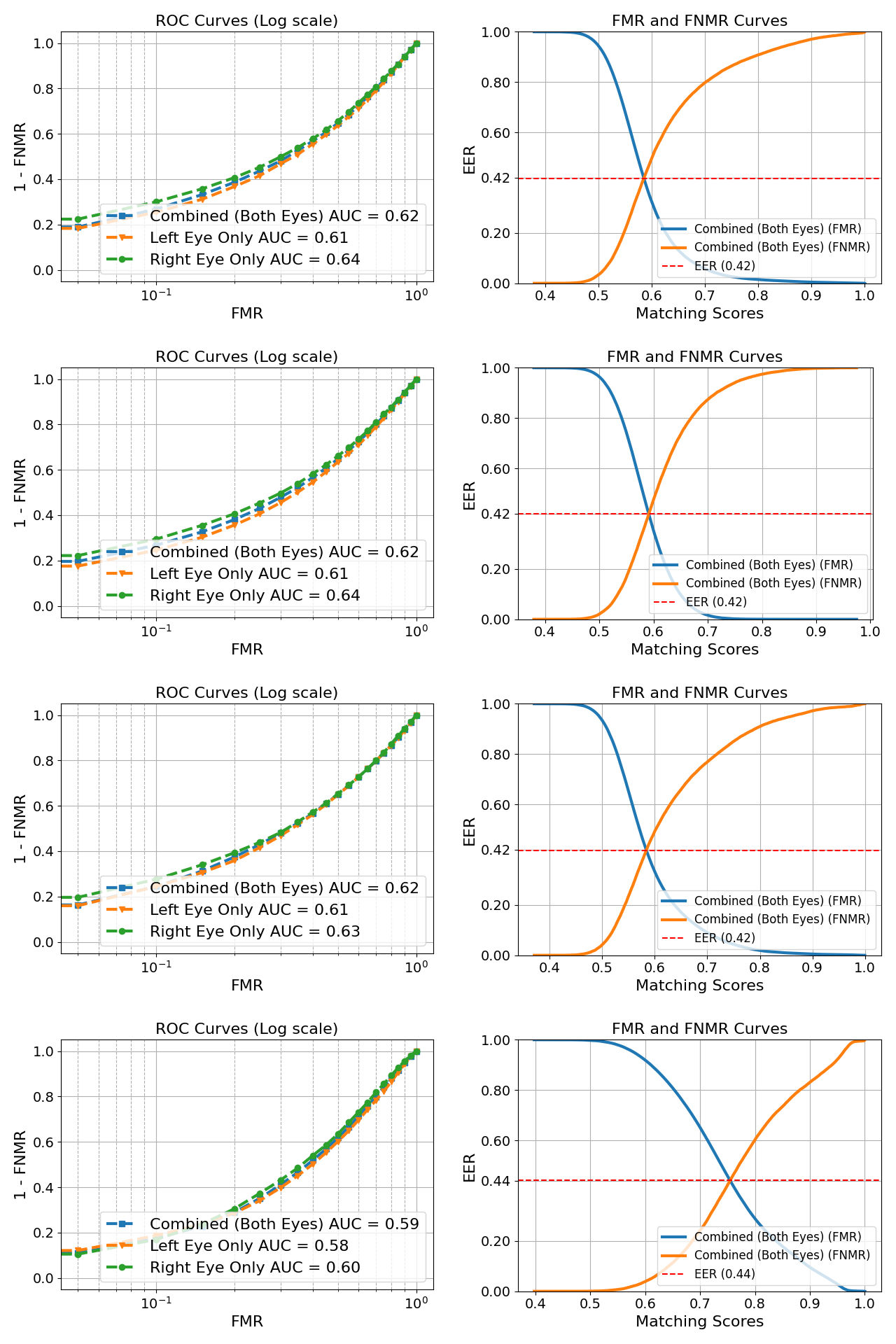}
    \caption{\textbf{Iris Recognition Performance:} Operational curves for unimodal iris verification across varying image adjustment pipelines. \textbf{Rows:} represent the four evaluation scenarios: (a) original, (b) off-axis correction (OAC), (c) specular glare removal (UnReflect), and (d) non-uniform illumination restoration (UNIR-Net). \textbf{Columns:} represent biometric metrics: the left column illustrates the log-scaled ROC curves, and the right column highlights the Equal Error Rate (EER) from the error curves.}
    \label{fig:degradation_matrix}
\end{figure*}

\begin{table}[t]
\centering
\footnotesize
\caption{\textbf{Periocular Recognition Performance:} Each cell reports \textbf{Original / UnReflect / UNIR-Net} processed VRBiom images, averaged over the three model folds\protect~\cite{Periocular} in terms of EER ($\downarrow$), AUC ($\uparrow$) and FNMR@FMR=$10^{-2}$ ($\downarrow$).}\label{tab2}
\begin{tabular}{|l|c|c|c|}
\hline
Model & EER (\%)  & AUC  & FNMR@FMR$=1e-2$  \\
\hline
MobileFaceNet & 34.70 / 31.97 / 35.14 & 0.74 / 0.76 / 0.73 & 0.69 / 0.70 / 0.71 \\
ResNet-50     & 36.37 / 32.32 / 36.54 & 0.72 / 0.76 / 0.70 & 0.67 / 0.68 / 0.71 \\
ResNet-18     & 35.83 / 31.65 / 36.47 & 0.73 / 0.76 / 0.70 & 0.67 / 0.68 / 0.71 \\
\hline
\end{tabular}
\end{table}

Fig.~\ref{fig:degradation_matrix} presents the Receiver Operating Characteristic (ROC) curves for unimodal iris recognition across our four evaluated image adjustment pipelines: (a) Original Input, (b) Off-Axis Correction (OAC), (c) Specular Glare Removal (UnReflect), and (d) Non-Uniform Illumination Restoration (UNIR-Net).

In the baseline (before image adjustment) ($1^{st}$ Row), unimodal iris verification exhibits high error rates, yielding an Area Under the Curve (AUC) of $0.62$ for both eyes combined ($0.61$ Left, $0.64$ Right). When evaluating geometric Off-Axis Correction ($2^{nd}$ Row), we observe similar performance ($\text{AUC} = 0.62$). This confirms that while projective homography matrices successfully restore circular proportions, the interpolation and boundary padding during perspective warping adversely affect the high-frequency texture. 

Our experiments expose an important \textit{limitation} of applying generative image restoration to biometric verification. While UnReflect ($3^{rd}$ Row) performs on par with the baseline ($\text{AUC} = 0.62$), the application of UNIR-Net ($4^{th}$ Row) results in a noticeable degradation of verification accuracy, dropping the combined AUC to $0.59$. This empirical finding indicates that although UNIR-Net successfully enhances macroscopic visual clarity and ISO contrast metrics, non-uniform illumination networks may hallucinate or smooth over the microscopic, non-redundant crypts and furrows of the iris texture that feature extractors rely upon for identity discrimination.

\begin{figure*}[t]
    \centering
    \includegraphics[width=0.6\textwidth]{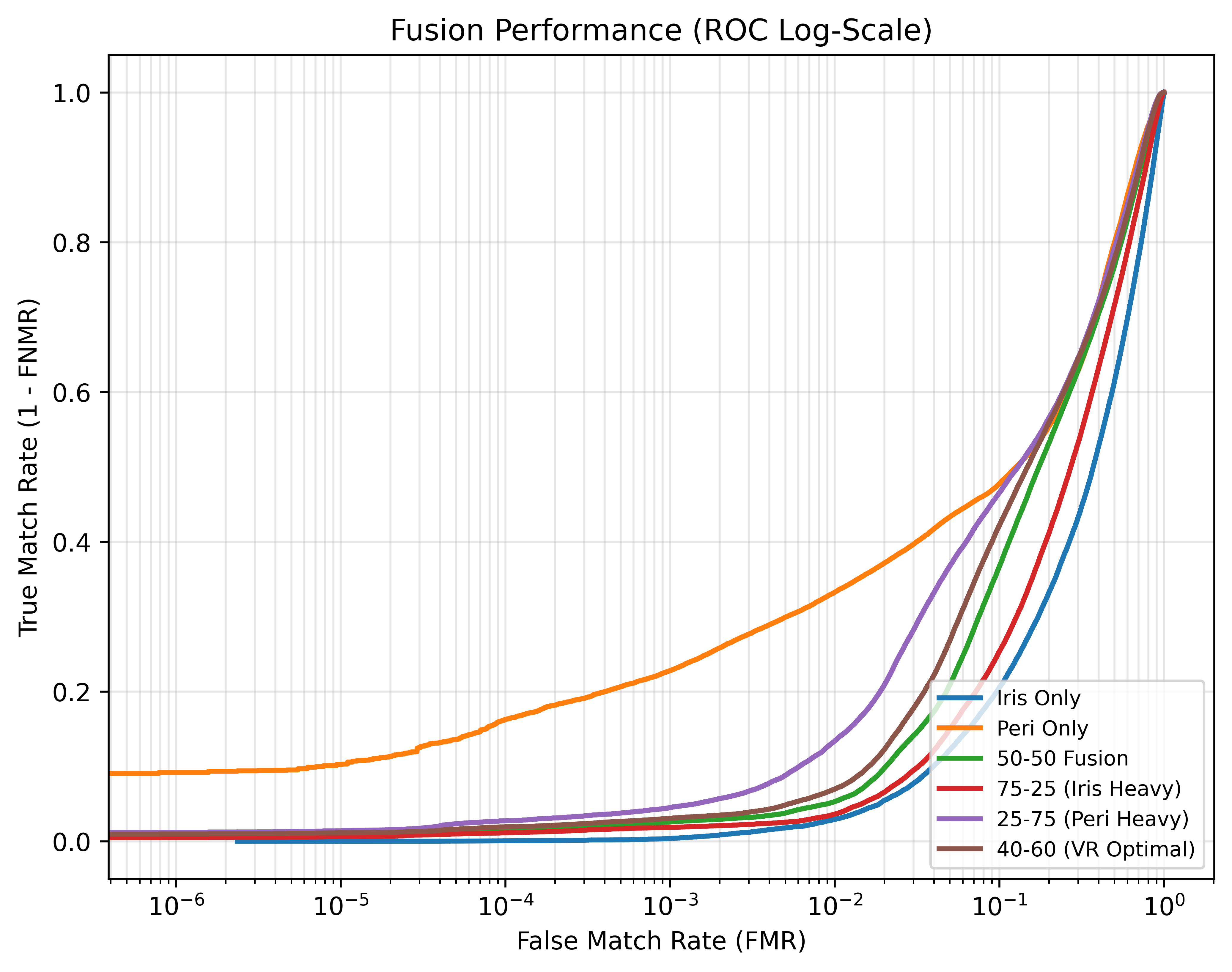}
    \caption{\textbf{Fusion analysis:} ROC curves comparing various fusion variations.}
    \label{fig:fusion_roc}
\end{figure*}

\textbf{Fusion.} Fig.~\ref{fig:fusion_roc} and Table \ref{tab:fusion_metrics} establish that periocular recognition alone ($\text{EER} = 0.34$, $\text{AUC} = 0.76$, $d' = 1.08$) substantially outperforms unimodal iris verification ($\text{EER} = 0.44$, $\text{AUC} = 0.59$, $d' = 0.34$). The performance gap indicates robustness of periocular features against corneal reflections and off-axis variations.

Weighted score-level fusion achieves performance gains that depend heavily on the modality weights; see Fig.~\ref{fig:fusion_roc}. An equal weighting strategy (Fusion 50/50: $w_i=w_p=0.5$) yields an EER of $0.34$ and a decidability index ($d'$) of $0.79$, effectively matching the periocular baseline but failing to leverage complementary gains due to the noise induced by the degraded iris channel. Over-reliance on iris features (Fusion 75/25: $w_i=0.75,w_p=0.25$) severely degrades system accuracy ($\text{EER} = 0.38$, $d' = 0.55$), resulting in higher false rejection.

Conversely, adopting a periocular-heavy weighting scheme improves recognition. We adopt the configuration from~\cite{Luiz} (Fusion 40/60: $w_i=0.40,w_p=0.60$) and it reduces the EER to $0.33$ while increasing $d'$ to $0.89$. The \textbf{best multimodal fusion verification accuracy} is achieved by the Fusion 25/75: $w_i=0.25,w_p=0.75$ (Peri-Heavy) configuration, which reduces the EER to $\textbf{0.33}$, increases the AUC to $0.75$, and achieves $\sim3\times$ improvement in  $\textbf{d' = 0.99}$ over unimodal iris recognition.

\section{Discussion and Future Work}
We presented a comprehensive analysis of near-infrared ocular data captured using VR head mounted displays in terms of ISO/IEC 29794-6 iris quality evaluation, ocular image adjustment, and verification using iris and periocular cues. Our analysis of the quality metrics on the VRBiom dataset exposes two critical limitations in current standardization: first, some metrics like margin adequacy and concentricity cannot be accommodated using the current settings; second, evaluation differs significantly based on the implementation due to variations in segmentation routines. Our investigation into ocular image adjustments reveals that although UnReflect used for specular glare removal systematically lowered the EER and elevated AUC across all periocular models (MobileFaceNet, ResNet-50, and ResNet-18), generative illumination restoration (UNIR-Net) and geometric off-axis correction failed to improve iris verification utility. We hypothesize that complex restoration networks can impair verification by over-smoothing fine iris patterns and may require domain adaptation. Our investigation into ocular recognition reveals that periocular recognition outperforms unimodal iris verification across unconstrained VR feeds. This disparity is likely due to the HMD camera hardware with inward-facing sensors that capture the eye at acute, off-axis angles resulting in strong perspective compression and occlusion by eyelids. Conversely, periocular features, including, canthi, eyebrows, and surrounding skin are less affected. Weighted score-level fusion improves the AUC by $\sim$16\% and lowers the EER by $\sim$11\% relative to unimodal iris recognition. Future work will focus on extending our analysis to more datasets and designing an end-to-end image restoration and multimodal verification pipeline with revised quality evaluation. Specifically, we aim to explore \textit{quality-driven adaptive fusion}, utilizing per-frame ISO metric scores to adjust the modality weights ($w_i, w_p$) instead of utilizing fixed parameters.

\begin{table}[t]
    \centering
    \footnotesize
    \caption{Performance evaluation metrics for \textit{unimodal} and \textit{multimodal} fusion schemes on the VRBiom dataset. Best values are \textbf{bolded}; best fusion* values are \underline{underlined}.}
    \label{tab:fusion_metrics}
    \begin{tabular}{lccc}
        \toprule
        \textbf{Configuration} & \textbf{EER} ($\downarrow$)& \textbf{AUC} ($\uparrow$) & \textbf{d'} ($\uparrow$)\\
        \midrule
        Iris Only & 0.44 & 0.59 & 0.34 \\
        Periocular Only & 0.34 & \textbf{0.76} & \textbf{1.08} \\
        Fusion 50/50 & 0.34 & 0.72 & 0.79 \\
        Fusion 75/25 (Iris Heavy) & 0.38 & 0.66 & 0.55 \\
        Fusion 25/75 (Peri Heavy)* & \underline{\textbf{0.33}} & \underline{0.75} & \underline{0.99} \\
        Fusion 40/60 (VR Optimal) & 0.33 & 0.73 & 0.89 \\
        \bottomrule
    \end{tabular}
\end{table}

\textbf{Acknowledgment.} This material is based upon work supported by the National Science Foundation REU Site under Grant No. 2349771. We would like to thank Prof. Adam Czajka for sharing the internal implementation of ISO/IEC 29794-6 quality metrics developed by the University of Notre Dame. We also thank the authors of the VRBiom dataset for releasing it for research.

%
%
\small
\bibliographystyle{splncs04}
\bibliography{mybibliography}

@misc{VRBiom,
      title={{VRBiom:} A New Periocular Dataset for Biometric Applications of HMD}, 
      author={Ketan Kotwal and Ibrahim Ulucan and Gokhan Ozbulak and Janani Selliah and Sebastien Marcel},
      year={2024},
      eprint={2407.02150},
      archivePrefix={arXiv},
      primaryClass={cs.CV},
      url={https://arxiv.org/abs/2407.02150}, 
}

@misc{CVRL,
      title={Lowering the Barrier to {IREX} Participation: Open-Source Algorithms, Toolkit, and Benchmarking for Iris Recognition}, 
      author={Siamul Karim Khan and Patrick J. Flynn and Adam Czajka},
      year={2026},
      eprint={2605.20735},
      archivePrefix={arXiv},
      primaryClass={cs.CV},
      url={https://arxiv.org/abs/2605.20735}, 
}

@article{Brazil,
author = {França, Rodrigo and Rosa, Renata and Rodriguez, Demostenes Zegarra},
year = {2020},
month = {07},
pages = {50471-50491},
title = {Iris image quality assessment based on {ISO/IEC 29794-6:2015} standard},
volume = {6},
journal = {Brazilian Journal of Development},
}

@misc{ImmerIris,
      title={ImmerIris: A Large-Scale Dataset and Benchmark for Off-Axis and Unconstrained Iris Recognition in Immersive Applications}, 
      author={Yuxi Mi and Qiuyang Yuan and Zhizhou Zhong and Xuan Zhao and Jiaogen Zhou and Fubao Zhu and Jihong Guan and Shuigeng Zhou},
      year={2026},
      eprint={2510.10113},
      archivePrefix={arXiv},
      primaryClass={cs.CV},
      url={https://arxiv.org/abs/2510.10113}, 
}

@INPROCEEDINGS{Periocular,
  author={Kolf, Jan Niklas and Boutros, Fadi and Kirchbuchner, Florian and Damer, Naser},
  booktitle={IEEE International Joint Conference on Biometrics (IJCB)}, 
  title={Lightweight Periocular Recognition through Low-bit Quantization}, 
  year={2022},
  volume={},
  number={},
  pages={1-12},
  }

@InProceedings{Rota_2026_CVPR,
    author    = {Rota, Alberto and Kiray, Mert and Karaoglu, Mert Asim and Ruhkamp, Patrick and De Momi, Elena and Navab, Nassir and Busam, Benjamin},
    title     = {{UnReflectAnything:} RGB-Only Highlight Removal by Rendering Synthetic Specular Supervision},
    booktitle = {Proceedings of the IEEE/CVF Conference on Computer Vision and Pattern Recognition (CVPR)},
    month     = {June},
    year      = {2026},
    pages     = {241-250}
}

@article{
Egocentric,
author = {Kuo Wang  and Ajay Kumar },
title = {Human Identification in Metaverse Using Egocentric Iris Recognition},
journal = {TechRxiv},
volume = {2022},
number = {0517},
pages = {},
year = {2022}}

@misc{Gaze,
      title={GazeShift: Unsupervised Gaze Estimation and Dataset for VR}, 
      author={Gil Shapira and Ishay Goldin and Evgeny Artyomov and Donghoon Kim and Yosi Keller and Niv Zehngut},
      year={2026},
      eprint={2603.07832},
      archivePrefix={arXiv},
      primaryClass={cs.CV},
      url={https://arxiv.org/abs/2603.07832}, 
}

@INPROCEEDINGS{5339068,
  author={Park, Unsang and Ross, Arun and Jain, Anil K.},
  booktitle={2009 IEEE 3rd International Conference on Biometrics: Theory, Applications, and Systems}, 
  title={Periocular biometrics in the visible spectrum: A feasibility study}, 
  year={2009},
  volume={},
  number={},
  pages={1-6},
 }

@INPROCEEDINGS{5597603,
  author={Woodard, Damon L. and Pundlik, Shrinivas and Miller, Philip and Jillela, Raghavender and Ross, Arun},
  booktitle={2010 20th International Conference on Pattern Recognition}, 
  title={On the Fusion of Periocular and Iris Biometrics in Non-ideal Imagery}, 
  year={2010},
  volume={},
  number={},
  pages={201-204},
  }

@article{KUMARI20221086,
title = {Periocular biometrics: A survey},
journal = {Journal of King Saud University - Computer and Information Sciences},
volume = {34},
number = {4},
pages = {1086-1097},
year = {2022},
issn = {1319-1578},
author = {Punam Kumari and K.R. Seeja}
}

@INPROCEEDINGS{7935110,
  author={Stokkenes, Martin and Ramachandra, Raghavendra and Raja, Kiran B. and Sigaard, Morten K. and Busch, Christoph},
  booktitle={2017 5th International Workshop on Biometrics and Forensics (IWBF)}, 
  title={Feature level fused templates for multi-biometric system on smartphones}, 
  year={2017},
  volume={},
  number={},
  pages={1-5},
  }

@ARTICLE{6508951,
  author={Tan, Chun-Wei and Kumar, Ajay},
  journal={IEEE Transactions on Image Processing}, 
  title={Towards Online Iris and Periocular Recognition Under Relaxed Imaging Constraints}, 
  year={2013},
  volume={22},
  number={10},
  pages={3751-3765},
  }

@ARTICLE{6915725,
  author={Mahalingam, Gayathri and Ricanek, Karl and Albert, A. Midori},
  journal={IEEE Transactions on Information Forensics and Security}, 
  title={Investigating the Periocular-Based Face Recognition Across Gender Transformation}, 
  year={2014},
  volume={9},
  number={12},
  pages={2180-2192},
  }

@article{daugman,
author = {Daugman, J.},
title = {How iris recognition works},
year = {2004},
issue_date = {January 2004},
publisher = {IEEE Press},
volume = {14},
number = {1},
issn = {1051-8215},
journal = {IEEE Trans. Cir. and Sys. for Video Technol.},
month = jan,
pages = {21–30},
numpages = {10}
}

@Article{electronics14020240,
AUTHOR = {Baek, Junho and Park, Yeongje and Seok, Chaelin and Lee, Eui Chul},
TITLE = {Noise-Robust Biometric Authentication Using Infrared Periocular Images Captured from a Head-Mounted Display},
JOURNAL = {Electronics},
VOLUME = {14},
YEAR = {2025},
NUMBER = {2},
ARTICLE-NUMBER = {240},
ISSN = {2079-9292},

}

@misc{simeoni2025dinov3,
      title={DINOv3}, 
      author={Oriane Siméoni and Huy V. Vo and Maximilian Seitzer and Federico Baldassarre and Maxime Oquab and Cijo Jose and Vasil Khalidov and Marc Szafraniec and Seungeun Yi and Michaël Ramamonjisoa and Francisco Massa and Daniel Haziza and Luca Wehrstedt and Jianyuan Wang and Timothée Darcet and Théo Moutakanni and Leonel Sentana and Claire Roberts and Andrea Vedaldi and Jamie Tolan and John Brandt and Camille Couprie and Julien Mairal and Hervé Jégou and Patrick Labatut and Piotr Bojanowski},
      year={2025},
      eprint={2508.10104},
      archivePrefix={arXiv},
      primaryClass={cs.CV},
      url={https://arxiv.org/abs/2508.10104}, 
}

@article{P_rez_Zarate_2025,
   title={{UNIR-Net:} A novel approach for restoring underwater images with non-uniform illumination using synthetic data},
   volume={163},
   ISSN={0262-8856},
   journal={Image and Vision Computing},
   publisher={Elsevier BV},
   author={Pérez-Zarate, Ezequiel and Liu, Chunxiao and Ramos-Soto, Oscar and Oliva, Diego and Pérez-Cisneros, Marco},
   year={2025},
   month=Nov, pages={105734} }

@article{DLSurvey,
author = {Hussein, Haval and Abduallah, Wafaa and Omer, Herman},
year = {2026},
month = {02},
pages = {},
title = {Iris Recognition Based Deep Learning: A Survey},
volume = {2},
journal = {Dasinya Journal for Engineering and Informatics},
}

@inproceedings{VREyeSAM,
author = {Sharma, Geetanjali and Nagaich, Dev and Jaswal, Gaurav and Nigam, Aditya and Ramachandra, Raghavendra},
year = {2026},
month = {01},
pages = {},
title = {VREyeSAM: Virtual Reality Non-Frontal Iris Segmentation using Foundational Model with uncertainty weighted loss},
}

@INPROCEEDINGS{Fusion,
  author={Boutros, Fadi and Damer, Naser and Raja, Kiran and Ramachandra, Raghavendra and Kirchbuchner, Florian and Kuijper, Arjan},
  booktitle={2020 IEEE 23rd International Conference on Information Fusion (FUSION)}, 
  title={Fusing Iris and Periocular Region for User Verification in Head Mounted Displays}, 
  year={2020},
  volume={},
  number={},
  pages={1-8},
  }

@ARTICLE{MSurvey,
  author={Wang, Yuntao and Su, Zhou and Zhang, Ning and Xing, Rui and Liu, Dongxiao and Luan, Tom H. and Shen, Xuemin},
  journal={IEEE Communications Surveys and Tutorials}, 
  title={A Survey on Metaverse: Fundamentals, Security, and Privacy}, 
  year={2023},
  volume={25},
  number={1},
  pages={319-352},}

@misc{Review,
      title={Biometrics in Extended Reality: A Review}, 
      author={Ayush Agarwal and Raghavendra Ramachandra and Sushma Venkatesh and S. R. Mahadeva Prasanna},
      year={2024},
      eprint={2411.10489},
      archivePrefix={arXiv},
      primaryClass={cs.CR},
      url={https://arxiv.org/abs/2411.10489}, 
}

@article{Luiz,
   title={Deep representations for cross‐spectral ocular biometrics},
   volume={9},
   ISSN={2047-4938},
   number={2},
   journal={IET Biometrics},
   publisher={Institution of Engineering and Technology (IET)},
   author={Zanlorensi, Luiz A. and Lucio, Diego Rafael and Britto Junior, Alceu de Souza and Proença, Hugo and Menotti, David},
   year={2020},
   month=Feb, pages={68–77} }

@article{Bowyer,
  title={Image understanding for iris biometrics: A survey},
  author={K. Bowyer and Karen Hollingsworth and Patrick J. Flynn},
  journal={Comput. Vis. Image Underst.},
  year={2008},
  volume={110},
  pages={281-307},

}

@Inbook{Wildes,
author="Wildes, Richard",
editor="Wayman, James
and Jain, Anil
and Maltoni, Davide
and Maio, Dario",
title="Iris Recognition",
bookTitle="Biometric Systems: Technology, Design and Performance Evaluation",
year="2005",
publisher="Springer London",
address="London",
pages="63--95",
isbn="978-1-84628-064-1"
}

@article{MA,
  title={Personal Identification Based on Iris Texture Analysis},
  author={Li Ma and Tieniu Tan and Yunhong Wang and Dexin Zhang},
  journal={IEEE Trans. Pattern Anal. Mach. Intell.},
  year={2003},
  volume={25},
  pages={1519-1533},
}

\end{document}